\documentclass{easychair}

\pdfoutput=1
\usepackage{tikz}
\usepackage{pgfplots}
\usepackage{todonotes}
\usepackage{cleveref}
\usepackage{hyperref}
\usepackage{pgfplotstable}
\pgfplotsset{compat=1.14}
\usepgfplotslibrary{groupplots}
\usepackage{microtype}

\pgfplotstableset{ integer column/.style={%
    /pgfplots/table/display columns/#1/.style={%
      numeric type,
    }
  },
  percentage column/.style={%
    /pgfplots/table/display columns/#1/.style={ precision = 1, fixed, multiply
      by = {100}, fixed zerofill,
    }
  }
}

\title{Tactic Learning and Proving for the Coq Proof Assistant\thanks{This work
    was supported by the \textit{AI4REASON} ERC Consolidator grant nr. 649043 and by the European Regional Development Fund under the project
    AI\&Reasoning (reg. no. CZ.02.1.01/0.0/0.0/15\_003/0000466)}}

\author{Lasse Blaauwbroek\inst{1, 2} \and Josef Urban\inst{1} \and Herman
  Geuvers\inst{2}}
\authorrunning{L. Blaauwbroek et al.}

\institute{Czech Technical University, Prague, Czech Republic \and Radboud
  University, Nijmegen, The Netherlands \\ \vspace{3pt}
  \email{lasse@blaauwbroek.eu, josef.urban@gmail.com, herman@cs.ru.nl}}

\titlerunning{Tactic Learning and Proving for the Coq Proof Assistant}

\makeatletter
\hypersetup{
  pdftitle={Tactic Learning and Proving for the Coq Proof Assistant},
  pdfsubject={Machine Learning for Theorem Provers},
  pdfauthor={Lasse Blaauwbroek, Josef Urban and Herman Geuvers},
  pdfkeywords={Coq Proof Assistant, Interactive Theorem Proving,
    Machine Learning, Tactic Search, Proof Synthesis}}
\makeatother

\begin{document}

\maketitle

\begin{abstract}
  We present a system that utilizes machine learning for tactic proof search in
  the Coq Proof Assistant. In a similar vein as the TacticToe project for HOL4,
  our system predicts appropriate tactics and finds proofs in the form of tactic
  scripts. To do this, it learns from previous tactic scripts and how they are
  applied to proof states. The performance of the system is evaluated on the Coq
  Standard Library. Currently, our predictor can identify the correct tactic to
  be applied to a proof state 23.4\% of the time. Our proof searcher can fully
  automatically prove 39.3\% of the lemmas. When combined with the CoqHammer
  system, the two systems together prove 56.7\% of the library's lemmas.
\end{abstract}

\section{Introduction}
The Coq Proof Assistant~\cite{the_coq_development_team_2019} is an Interactive
Theorem Prover in which one proves lemmas using a tactic language. This language
contains a wide range of tactics, from simple actions like \verb|intros| and
\verb|apply| to complicated decision procedures such as \verb|ring| and
\verb|tauto| and search procedures like \verb|auto| and \verb|firstorder|.
Although the second and third options provide a large amount of automation, they
still require human intervention to use. (1) Decision procedures only apply to
particular domains, requiring the user to know when they are appropriate, and
(2) Coq's search tactics require careful construction of hint databases to be
performant.

The system we present in this paper learns from previously written proof scripts
and how they are applied to proof states. This knowledge can then be used to
suggest a tactic to apply on a previously unseen proof state or to perform a
full proof search and prove the current lemma automatically. Learning on the
tactic level has some advantages over learning from low-level proof terms in
Gallina, Coq's version of the Calculus of Inductive
Constructions~\cite{DBLP:conf/tlca/Paulin-Mohring93}: (1) Tactics represent
coarser proof steps than the individual identifiers in a proof term. They are
also more forgiving; a minor modification to a tactic script is more likely to
be nondestructive than a minor modification to a proof term, making the machine
learning task easier. (2) By learning on the tactic level, we allow the user to
introduce domain-specific knowledge to the system by writing custom tactics. By
using these tactics in hand-written proofs, the system will automatically learn
of their existence and start to use them.

Similar to the TacticToe project for HOL4~\cite{DBLP:conf/lpar/GauthierKU17},
our system has the following components.
\begin{description}
\item[Proof recording] First, we create a database of tactics that are executed
  in existing (human-written) tactic scripts, recording in which proof states
  they are executed.
\item[Tactic prediction] Using this database of pairs of tactics and proof
  states, we use machine learning to generate a list of tactics likely
  applicable to the current proof state. This list can either be used as a
  tactic recommendation system for the user, or as input to the next component.
\item[Proof search] Finally, for a given proof state, we perform a proof search
  by repeatedly applying predicted tactics to the proof state and subsequent
  states in the search tree until all proof obligations are discharged.
\end{description}
These components are all implemented in OCaml (the implementation language of
Coq)---without external dependencies---and integrated directly with Coq,
enabling us to eventually create a user-facing tool that is easy to use,
install, and maintain.

In this paper, we present the technical implementation of the components above
and an evaluation of the quality of tactic predictions and proof search. In
\Cref{sec:related-work}, we give an overview of related work. Then, in
\Cref{sec:recording,sec:tactic-prediction,sec:proof-search}, we take a deeper
dive into the different components of our system. Finally, \Cref{sec:evaluation}
contains an evaluation of our system on the Coq Standard Library. The dataset
used for our evaluation, together with the software that generated it is
publicly available~\cite{blaauwbroek_lasse_2020_504435}.

\section{Related Work}
\label{sec:related-work}

The high level of integration of our system with Coq is a distinguishing feature
compared to existing learning-guided systems for Coq. The ML4PG
system~\cite{DBLP:journals/corr/abs-1212-3618} provides tactic predictions and
other statistics but is instead integrated with the Proof
General~\cite{DBLP:conf/tacas/Aspinall00} proof editor and requires connections
to Matlab or Weka to function. The SEPIA
system~\cite{DBLP:conf/cade/GransdenWR15} provides proof search using tactic
predictions and is also integrated with Proof General. Note, however, that its
proof search is only based on tactic traces and does not make predictions based
on the proof state. GamePad~\cite{DBLP:conf/iclr/HuangDSS19} is a framework that
integrates with the Coq codebase and allows machine learning to be performed in
Python. We are not aware of a comprehensive evaluation of tactic prediction in
this system. Finally, CoqGym~\cite{DBLP:conf/icml/YangD19} also extracts tactic
and proof state information on a large scale and uses it to construct a deep
learning model capable of generating full tactic scripts.

The main inspiration for our system is the
TacticToe~\cite{DBLP:conf/lpar/GauthierKU17} system for
HOL4~\cite{DBLP:conf/tphol/SlindN08}. Our work is similar both in doing a
learning-guided tactic search and by its complete integration in the underlying
proof assistant without relying on external machine learning libraries and
specialized hardware. More recent TacticToe-inspired systems that do not aim at
complete independence of external learning toolkits include
HOList~\cite{DBLP:conf/icml/BansalLRSW19} and similar work by Wu et al. for
HOL4~\cite{Wu2020}.

Significant work on learning-based guidance has been done in the context of
connection-based and saturation-based Automated Theorem Provers (ATPs) such as
leanCoP~\cite{DBLP:journals/jsc/OttenB03} and E~\cite{DBLP:conf/lpar/Schulz13}.
In the leanCoP setting, (reinforcement) learning of clausal steps based on the
proof state is done in systems such as
rlCoP~\cite{DBLP:conf/nips/KaliszykUMO18,DBLP:journals/corr/abs-1911-12073}, and
(FE)MaLeCoP~\cite{DBLP:conf/lpar/KaliszykU15,DBLP:conf/tableaux/UrbanVS11}. In
E, given-clause selection has been guided by learning in systems such as ENIGMA
and its
variants~\cite{DBLP:conf/mkm/JakubuvU17,DBLP:conf/lpar/LoosISK17,DBLP:conf/mkm/JakubuvU18,DBLP:conf/cade/ChvalovskyJ0U19,DBLP:conf/tableaux/GoertzelJU19}.

ATPs are also used in
\emph{hammers}~\cite{DBLP:journals/jfrea/BlanchetteKPU16,DBLP:journals/jar/KaliszykU14,DBLP:journals/jar/BlanchetteGKKU16,DBLP:journals/jar/KaliszykU15a},
where learning-based premise selection is used to translate an ITP problem to
the ATP's target language (typically First Order Logic). A proof found by the
ATP can then be reconstructed in the proof assistant. A particular hammer
instance for Coq is the CoqHammer system~\cite{DBLP:journals/jar/CzajkaK18}.

\section{Proof Recording}
\label{sec:recording}

The first component of the system is the recording of existing proofs. As said,
this is done on the level of tactics. Conceptually speaking, when a tactic
script is executed, we record the proof state before and after the execution of
each tactic. The difference between these two states represents the action
performed by the tactic, while the state before the tactic represents the
context in which it was useful. By recording many such instances for a tactic,
we create a dataset representing an approximation of the semantic meaning of
that tactic. In practice, we currently only record the proof state before the
execution of a tactic because, during proof search, we do not yet evaluate the
proof state's quality after the execution of a tactic.

An important question is what exactly constitutes a tactic in Coq. On a low
level, Coq utilizes a proof monad in which tactics can be written on the OCaml
level~\cite{DBLP:journals/jlp/KirchnerM10}. This monad supports backtracking,
among other things. It is, however, not immediately accessible by end-users. For
that, several tactic languages exist that are interpreted into the proof
monad~\cite{DBLP:journals/pacmpl/KaiserZKRD18,DBLP:conf/esop/MalechaB16,DBLP:journals/jfrea/GonthierM10,pedrot2019ltac2}.
The most used language is Ltac1~\cite{DBLP:conf/lpar/Delahaye00}. In an ideal
world, we would record tactics on the level of the proof monad since that would
allow us to record tactics from all existing tactic languages. However, it does
not appear to be possible to put recording instrumentation in place on the
monadic level. Therefore, we have chosen to only record tactics from the Ltac1
language.

Within the Ltac language, it is also not immediately clear what a tactic is. One
option is to decompose a script into a series of the most primitive tactic
invocations and record those. On the other side of the extreme, one could view
every whole vernacular expression as one tactic. The first option greatly
diminishes the advantages of the system because such low-level recording would
not allow domain-specific decision procedures or custom tactics to be recorded.
The second option means that many tactics will be unique and rather specific to
a single proof state. A good solution will lie somewhere in between. Striking
this balance is further complicated by the possibility of tactics to backtrack.
It is unclear whether we should record only the successful trace of a tactic, or
keep the backtracking procedure as one building block.

At the moment, we see every vernacular command as one tactic, except for tactic
composition (\verb|tac1; tac2|) and tactic dispatching (\texttt{tac1; [tac2 |
  tac3]}). To record a tactic script, conceptually, we replace a tactic \verb|t|
with a custom recording tactic \verb|r t| that receives the original tactic as
an argument. Tactic \verb|r t| first records the current proof state before
executing $\verb|t|$. Then this proof state is added to the database together
with tactic \verb|t|. As an example of a tactic decomposition and recording, the
tactic expression \verb=tac1; [tac2 | tac3]; tac4= will be converted to
\verb=r tac1; [r tac2 | r tac3]; r tac4=.

\section{Tactic Prediction}
\label{sec:tactic-prediction}

After creating a database of tactics and recorded proof states, we want to
predict the correct tactic to make progress in a new, unseen proof state. For
this, we must find a good characterization of the sentences present in a proof
state. Currently, this characterization is a set of features. The features of a
sentence are all of its one-shingles and two-shingles, meaning they are all
identifiers and pairs of adjacent identifiers in the abstract syntax tree. For
example, the sentence $(x + y) + z = x + (y + z)$ will be converted to the
feature set
\[\{eq, plus, x, y, z, eq\_plus, plus\_x, plus\_y, plus\_z, eq\_z, eq\_x\}.\]
For the characterization of a proof state, we take the union of the feature set
of all sentences in its hypotheses and the goal.

To find a list of likely matches for a new proof state, we run a $k$-nearest
neighbor algorithm on the vectors. We compare vectors using several standard
metrics and similarities (specialized for feature sets instead of feature
vectors):
\[\text{cosine}(f_1, f_2) = \frac{|f_1\cap f_2|}{\sqrt{|f_1||f_2|}}\quad
  \text{euclid}(f_1,f_2) = \sqrt{|f_1 \cup f_2| - |f_1\cap f_2|}\quad
  \text{jaccard}(f_1,f_2) = \frac{|f_1\cap f_2|}{|f_1\cup f_2|}
\]
Finally, we would like to weigh features based on their relative importance in a
given proof state with respect to the entire database. For this, we use the
TfIdf statistic~\cite{DBLP:journals/jd/Jones04} (which is the same as the Idf
statistic since we use feature sets rather than bags). We then use a generalized
version of the Jaccard index to incorporate these weights into the predictions.
\[\text{tfidf}(x) = \log \frac{N}{|x|_N} \quad\quad\quad\quad\quad
  \text{jaccard}_w(f_1,f_2) = \frac{\sum_{x \in f_1\cap
      f_2}\text{tfidf}(x)}{\sum_{x\in f_1\cup f2}\text{tfidf}(x)}\] Here $N$ is
the database size, and $|x|_N$ is the number of times feature $x$ occurs in the
database.

Our $k$-nearest neighbor algorithm outputs an ordered list $l$ of pairs $\langle
(d_i, t_i)\rangle_k$, where $t_i$ is a tactic and $d_i$ the similarity between
the current proof state and the proof state on which $t_i$ was applied. If $l$
contains duplicate tactics, we select only the pair with the best score. The
list is ordered by the similarities d and can, therefore, immediately be used as
a list of tactic recommendations.

A major downside of the $k$-NN approach above is that it requires an exhaustive
search of the entire tactic database to obtain a list of recommendations.
Therefore, the time complexities of the algorithms grow linearly with the size
of the database, which can be quite large. To eliminate slow queries, we include
an approximate $k$-NN algorithm that employs Locally Sensitive
Hashing~\cite{DBLP:conf/vldb/GionisIM99}. LSH is a technique that uses hashing
functions that map similar items to the same integer with high probability (as
opposed to the normal hashing function in which collisions are avoided). These
functions are then used with hashtables to map similar items to the same bucket,
at which point they can be retrieved in constant time. As long as the notion of
similarity of the hashing function agrees with the intended similarity, this can
be used as a substitute for the corresponding exhaustive $k$-NN algorithm. The
downside of this method is that it is approximate and can thus miss crucial
tactics. In our implementation, we use a technique called LSH
Forest~\cite{DBLP:conf/www/BawaCG05} to approximate the Jaccard index described
above.

\section{Proof Search}
\label{sec:proof-search}

It is unlikely that the tactic prediction system will predict the correct tactic
every time. For this reason, a proof search must be performed, where not only
the first predicted tactic is chosen. We recursively explore the proof tree by
executing a predicted tactic and then predicting a new list of tactics based on
the new proof state. We explore this tree using a skewed form of breadth-first
search, which we call diagonal search. Given a list of predictions, $\langle
(d_i, t_i)\rangle_k$, we always explore the subtree starting from tactic $t_i$
one layer deeper than the subtree starting from tactic $t_{i+1}$. This diagonal
exploration is continued recursively.

Diagonal search captures the idea that tactics that are predicted as better
should be explored more vigorously than other tactics. In the future, we intend
to refine this search style by using the similarity scores $d_i$ to determine
how ``diagonal'' the search should be. Another interesting approach is to take
inspiration from AlphaGo Zero~\cite{silver2017mastering} and use a Monte Carlo
Tree Search algorithm for proof search. We should note, however, that such
strategies have not born much fruit for the TacticToe system of HOL4.

\section{Evaluation}
\label{sec:evaluation}

We use Coq's standard library to evaluate the effectiveness of our system. The
standard library is an impartial choice that should represent the de facto proof
script style (although we acknowledge that many proof styles are available to
users). The library contains modules with a variety of complexity. Basic modules
regarding logic and arithmetic contain relatively simple proofs, while more
complicated developments like the reals are also included. Evaluating the
standard library also allows us to compare our performance to CoqHammer. The
dataset for the evaluation and the software that generated it is publicly
available~\cite{blaauwbroek_lasse_2020_504435}.

\subsection{Methodology}
\label{sec:methodology}

For our evaluation, we try to simulate how a hypothetical user might develop the
standard library. We assume that when the user tries to prove a new lemma, all
previous lemmas are already proven and can be used for learning. Hence, when
evaluating a Coq source code file, the tactic database starts with only
information inherited from the file's dependencies, if any. When we evaluate a
lemma in the file, we are only allowed to use state-tactic pairs from dependency
files and pairs recorded in the file's previous lemmas. In contrast, Hammer
evaluations typically use the less faithful approach of sorting the files
topologically and letting all previous proofs (not just those in the
dependencies) inform the current search, giving them more learning data.

We start with an offline evaluation of the feature quality and tactic prediction
quality. Then follows a full evaluation of the search strategy and a comparison
to CoqHammer. The standard library of Coq v8.10 is used in the evaluations. All
the experiments were run on a 32 cores Intel(R) Xeon(R) CPU E5-2698 v3 @ 2.30GHz
server with 256 GB memory.

\subsection{Tactic Prediction Evaluation}
\label{sec:feature-evaluation}

The quality of the tactic prediction directly influences the ability to
synthesize entire proofs. As the tactic predictions improve, the proof search
procedure will require less backtracking. To judge the prediction quality, we
perform an offline evaluation. We have extracted a textual representation of the
tactic database of every file. For every pair of proof state $s$ and tactic $t$
in a file, we use the previous pairs in that file and its dependencies to
predict an ordered list of candidate tactics. We judge this prediction by how
close to the top of the candidate list $t$ occurs.

An interesting question is whether a prediction may use tactics that have been
recorded in the lemma currently being proved. If so, the system could learn from
a proof while the user writes it, letting the user benefit from this within the
same proof. A problem with this is that the proof states within the same lemma
are likely to be very similar, especially when using our feature
characterization. Therefore, if we allow intra-lemma learning, we can expect to
see the tactics from the current lemma predicted often.

\begin{figure}
  \centering
  \begin{tikzpicture}
    \begin{groupplot}[ group style = {group size = 2 by 1}, ymin=0, ymax=.79,
      xmin=0, xmax=30, legend columns=-1, legend
      style={at={(-45pt,-25pt)},anchor=west}, height=185pt, ]
      
      \nextgroupplot[ extra y ticks = {0.7675912509487338}, width=0.5\textwidth,
      ylabel = {Cumulative Frequency}, title={$k$-NN with intra-lemma learning},
      ]
        
      \addplot[black] table [x expr=\thisrowno{0}, y expr=\thisrowno{1}, col
      sep=space] {./data/cumulative200/Cosine.noflush.deps.eval};
      \addplot[black] table [x expr=\thisrowno{0}, y expr=\thisrowno{1}, col
      sep=space] {./data/cumulative200/Euclid.noflush.deps.eval};
      \addplot[black] table [x expr=\thisrowno{0}, y expr=\thisrowno{1}, col
      sep=space] {./data/cumulative200/Jaccard.noflush.deps.eval};
      \addplot[black] table [x expr=\thisrowno{0}, y expr=\thisrowno{1}, col
      sep=space] {./data/cumulative200/TFIDF.noflush.deps.eval};
      \addplot[mark=|] table [x expr=\thisrowno{0}, y expr=\thisrowno{1}, col
      sep=space] {./data/cumulative200/LSHForest.noflush.deps.eval};
      \addplot[dotted] table [x expr=\thisrowno{0}, y expr=\thisrowno{1}, col
      sep=space] {./data/cumulative200/Random.noflush.deps.eval};
      \addplot[dashed] table [x expr=\thisrowno{0}, y expr=\thisrowno{1}, col
      sep=space] {./data/cumulative200/ReturnWholeDb.noflush.deps.eval};
      \addplot[dash dot dot] coordinates {(0,0.767591250948)
        (30,0.767591250948)};
      
      \legend{Cosine,Euclid,Jaccard,TfIdf,LSHF,Random,Reverse,Max}
      
      \nextgroupplot[ extra y ticks = {0.7033050438142552}, width=0.5\textwidth,
      title={$k$-NN without intra-lemma learning}, ]

      \addplot[black] table [x expr=\thisrowno{0}, y expr=\thisrowno{1}, col
      sep=space]{./data/cumulative200/Cosine.flush.deps.eval}; \addplot[black]
      table [x expr=\thisrowno{0}, y expr=\thisrowno{1}, col
      sep=space]{./data/cumulative200/Euclid.flush.deps.eval}; \addplot[black]
      table [x expr=\thisrowno{0}, y expr=\thisrowno{1}, col
      sep=space]{./data/cumulative200/Jaccard.flush.deps.eval}; \addplot[black]
      table [x expr=\thisrowno{0}, y expr=\thisrowno{1}, col
      sep=space]{./data/cumulative200/TFIDF.flush.deps.eval}; \addplot[mark=|]
      table [x expr=\thisrowno{0}, y expr=\thisrowno{1}, col
      sep=space]{./data/cumulative200/LSHForest.flush.deps.eval};
      \addplot[dotted] table [x expr=\thisrowno{0}, y expr=\thisrowno{1}, col
      sep=space]{./data/cumulative200/Random.flush.deps.eval}; \addplot[dashed]
      table [x expr=\thisrowno{0}, y expr=\thisrowno{1}, col
      sep=space]{./data/cumulative200/ReturnWholeDb.flush.deps.eval};
      \addplot[dash dot dot] coordinates {(0,0.70330504) (30,0.70330504)};
      
    \end{groupplot}
  \end{tikzpicture}
  \caption{Evaluation of different metrics used for $k$-nearest neighbor
    prediction. For every $k$, the proportion of state-tactic pairs for which
    the correct tactic is in the candidate list is plotted.}
  \label{fig:nearest-neighbor}
\end{figure}
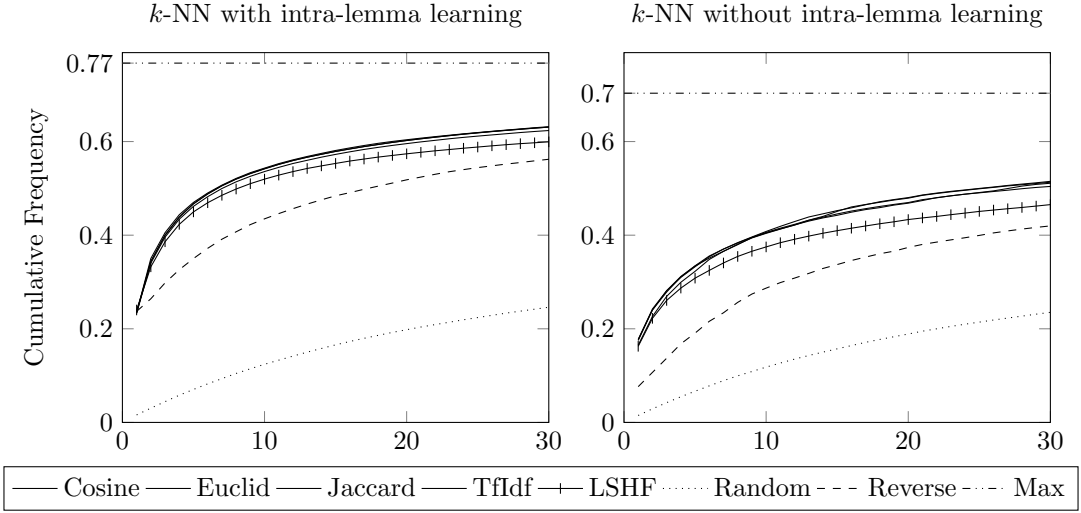

We evaluate the different metrics we proposed for the $k$-nearest neighbor
algorithm with and without the intra-lemma learning described above. The results
are shown in \Cref{fig:nearest-neighbor}. It turns out that all our metrics
perform quite similarly. Contrary to prior expectations, even the TfIdf-weighted
version of the Jaccard measure does not improve the predictions. To evaluate the
level of locality enjoyed by tactics, we also include the ``reverse-database''
predictor, where we rank tactics by when they were last added to the database.
When we allow intra-lemma learning, reverse-database does not perform much worse
than the other predictors, suggesting that predictors frequently output the
previous tactic as their best choice. Obviously, this will seldom be the correct
tactic. A future line of research is to incorporate the local tactic execution
history ({\emph memory}) into the predictors to combat this problem. Note,
however, that with a larger prediction pool, the reverse-predictor's performance
starts to go down, showing that our real predictors do nontrivial learning.
Finally, the LSHF predictor is performing only about 15\% worse than the rest.
Given that it performs predictions in constant time, allowing for approximately
a thousand predictions per second, this is very promising. Especially
considering that the file with the largest database contains 90594 tactics,
where an exhaustive prediction can take hundreds of milliseconds.

The main result of this evaluation is that the first predicted tactic is the
correct one 23.4\% of the time with intra-lemma learning and 17.7\% without
intra-lemma learning. This number goes up as we start to consider tactics that
are predicted with lower similarity. For $k=10$, the correct tactic is predicted
54.3\% and 40.3\% of the time, respectively. These results are encouraging,
considering that the theoretical best we can achieve is 77\% because the correct
tactic is not always in the database and thus not predictable.

\subsection{Proof Search Evaluation}
\label{sec:standard-library-evaluation}

\begin{table}
  \centering { \setlength{\tabcolsep}{4.8pt} \begin{filecontents*}{data/search-eval-table.csv}
development	lemmas	median	mad	max	1000deps	5000deps	10000deps	20000deps	deps	lshf	nodeps	2000deps	total
Arith	293	2.00000	5.00000	42	0.48805	0.51195	0.51195	0.51195	0.51195	0.61775	0.38225	0.52218	0.67235
Bool	130	1.00000	3.00000	8	0.93077	0.93077	0.92308	0.92308	0.93077	0.93846	0.83846	0.92308	0.93846
Classes	191	1.00000	4.00000	19	0.79581	0.78010	0.78534	0.80105	0.80105	0.80105	0.72251	0.78010	0.82723
FSets	1137	6.00000	15.00000	471	0.29639	0.28408	0.28672	0.27265	0.27704	0.32454	0.27617	0.28936	0.36236
Init	166	3.00000	6.00000	257	0.71084	0.71687	0.71687	0.71687	0.71687	0.68675	0.62048	0.71687	0.72892
Lists	388	8.00000	15.00000	74	0.36598	0.34278	0.34021	0.34278	0.34278	0.40464	0.36082	0.35825	0.43814
Logic	329	4.00000	9.00000	121	0.29787	0.29787	0.29787	0.29179	0.29483	0.33739	0.24316	0.29483	0.34954
MSets	830	5.00000	15.00000	164	0.36145	0.34819	0.34337	0.33614	0.33494	0.36867	0.34699	0.34578	0.42289
NArith	288	5.00000	10.00000	54	0.36111	0.35417	0.34722	0.34375	0.34375	0.38542	0.34722	0.37847	0.45486
Numbers	2198	5.00000	11.00000	275	0.16106	0.15833	0.15696	0.15332	0.15469	0.16879	0.14604	0.16333	0.20064
PArith	280	4.00000	9.00000	166	0.31786	0.32857	0.32500	0.32857	0.31071	0.30357	0.29643	0.32143	0.36786
Program	28	2.00000	5.00000	11	0.82143	0.78571	0.78571	0.78571	0.78571	0.71429	0.42857	0.78571	0.82143
QArith	295	5.00000	9.00000	80	0.34576	0.37288	0.34576	0.33220	0.33220	0.38305	0.31525	0.39322	0.47458
Reals	1975	8.00000	21.00000	1121	0.24253	0.23038	0.20962	0.19797	0.18278	0.24962	0.23241	0.23443	0.31443
Relations	37	8.00000	11.00000	25	0.32432	0.29730	0.29730	0.29730	0.29730	0.32432	0.32432	0.29730	0.37838
Setoids	4	3.50000	5.00000	5	1.00000	1.00000	1.00000	1.00000	1.00000	1.00000	0.00000	1.00000	1.00000
Sets	222	6.00000	12.00000	93	0.40991	0.38739	0.38739	0.39189	0.39189	0.48198	0.33333	0.40541	0.51351
Sorting	136	7.50000	15.00000	100	0.25735	0.21324	0.21324	0.19118	0.19118	0.24265	0.23529	0.24265	0.32353
Strings	74	9.00000	117.00000	1539	0.17568	0.21622	0.27027	0.24324	0.24324	0.21622	0.13514	0.18919	0.29730
Structures	390	3.00000	7.00000	71	0.40769	0.39744	0.39744	0.39487	0.39231	0.50000	0.33333	0.39231	0.54103
Vectors	37	7.00000	12.00000	22	0.18919	0.18919	0.18919	0.18919	0.18919	0.37838	0.21622	0.24324	0.43243
Wellfounded	36	8.00000	22.00000	49	0.19444	0.11111	0.11111	0.11111	0.11111	0.16667	0.08333	0.11111	0.19444
ZArith	952	3.00000	7.00000	87	0.46534	0.47059	0.45273	0.45063	0.44958	0.47584	0.35399	0.47479	0.57983
Total	10416	5.00000	12.00000	1539	0.31999	0.31404	0.30732	0.30136	0.29868	0.34044	0.28399	0.31893	0.39257
\end{filecontents*}
\pgfplotstabletypeset[ col sep=tab,
    columns={development,lemmas,median,mad,max,nodeps,1000deps,2000deps,10000deps,deps,lshf,total},
    every head row/.append style={ before row = {&& \multicolumn{3}{c|}{Proof
          Length}& \multicolumn{5}{c|}{Exhaustive $k$-NN with db size $n$}
        &&\\}, after row = {\hline} }, integer column/.list={1,4}, percentage
    column/.list={5,6,7,8,9,10}, columns/development/.append style={ string
      type, column name = {Development}, column type={r|} },
    columns/lemmas/.append style={column name = {\#Lem}, column type = r|},
    columns/median/.append style={column name = {$Q_2$}, column type = c,
      precision=0, fixed}, columns/mad/.append style={column name = {$Q_3$},
      column type = r, precision=0, fixed}, columns/max/.append style={column
      name = {Max}, column type = r|, precision=1, fixed},
    columns/nodeps/.append style={column name = {File}, column type = r},
    columns/1000deps/.append style={column name = {1000}, column type = r},
    columns/2000deps/.append style={column name = {2000}, column type = r},
    columns/10000deps/.append style={column name = {10000}, column type = r},
    columns/deps/.append style={column name = {All}, column type = r},
    columns/lshf/.append style={column name = {LSHF}, column type = |r},
    columns/total/.append style={ column name = {Total}, column type = {|r},
      precision = 1, fixed, multiply by = {100}, fixed zerofill }, every last
    row/.append style={before row = {\hline}}, ] {data/search-eval-table.csv}}
  \caption{A breakdown of the performance of some configurations of the system
    against the developments in the standard library. Columns $Q_2$ and $Q_3$
    are the second and third quartiles.}
  \label{tbl:development-breakdown}
\end{table}

For a full evaluation of the system, we want to know for how many of the lemmas
in the library a proof can be synthesized automatically. These experiments are
performed as follows. Every Coq source code file is compiled as usual. However,
when the system encounters a lemma, it disregards the original proof and tries
to synthesize a new proof based on the current tactic database. It then records
whether the proof synthesis was successful and substitutes the original proof
back, thus ensuring that it can progress through the entire file in case no
proof is found. The tactics used in the original proofs are then added to the
tactic database. Each proof search is given 40 seconds, equal to the total time
CoqHammer used in its evaluation.

We experiment with several configurations of our system. Exhaustive $k$-NN
search is performed using the TfIdf-weighted Jaccard measure. To keep the search
speed under control, we limit the size of the tactic database. First, we
evaluate with only tactics that have executed in the current file (no tactics
from dependencies). Then we evaluate using only the last 1000, 2000 and 10000
recorded tactics at the moment a proof search is initiated. For completeness, we
also evaluate using all available tactics. Finally, an evaluation using the
approximate Jaccard predictor LSHF is performed.

\Cref{tbl:development-breakdown} gives an overview of the performance of these
evaluations, broken down for the different developments of the standard library.
The different configurations can prove between 28.4\% and 34.0\% of the library
lemmas, with the LSHF configuration being the best. Together, the configurations
prove 39.3\% of the library lemmas. From this, it is clear that the most crucial
factor is the speed at which the tactic predictions can be made. The faster the
predictions, the deeper the search tree can be explored. Since LSHF's speed is
in the order of a thousand predictions per second when using the full tactic
database, it defeats the brute force $k$-NN approaches despite missing some
tactics due to its approximate nature. For the exhaustive configurations, the
versions with smaller databases work the best. Therefore, we conclude that
tactics exhibit good locality. That is, tactics that were recently executed are
likely to be useful again. It should be noted that the exhaustive configuration
with a full database still proves lemmas that are not proven by any other
configuration. In those cases, its slow proof exploration is compensated by the
fact that it can find exactly the right tactic.

When looking at the different developments in the standard library, we see
substantial differences in performance, mostly due to differences in proof
styles, with longer and shorter proofs. For example, in \verb|Arith|, we do
quite well due to short proofs facilitated by existing automation, such as
\verb|auto|. The most problematic development is \verb|Numbers| due to its size
and low success rate caused by longer proofs and complicated tactics that are
difficult to decompose.

\begin{figure}
  \centering
  \begin{tikzpicture}
    \begin{semilogyaxis}[ xlabel={Proof length}, ylabel={\#Occurrences},
      xmax=80, log origin=infty, enlargelimits=false, width=\textwidth,
      height=119pt]
      \addplot[fill=gray, color=gray] table [x, y, col sep=comma]
      {data/original-lengths.txt} \closedcycle;
      \addplot[fill] table [x, y, col sep=comma] {data/proof-lengths.txt} \closedcycle;
      \addplot[dashed, lightgray] coordinates {(0,100) (80,100)};
      \addplot[dashed, lightgray] coordinates {(0,10) (80,10)};
      \addplot[dashed, lightgray] coordinates {(0,1000) (80,1000)};
    \end{semilogyaxis}
  \end{tikzpicture}
  \caption{Lengths of successfully synthesized proofs vs. the original proof
    lengths in the library.}
  \label{fig:proof-lengths}
\end{figure}
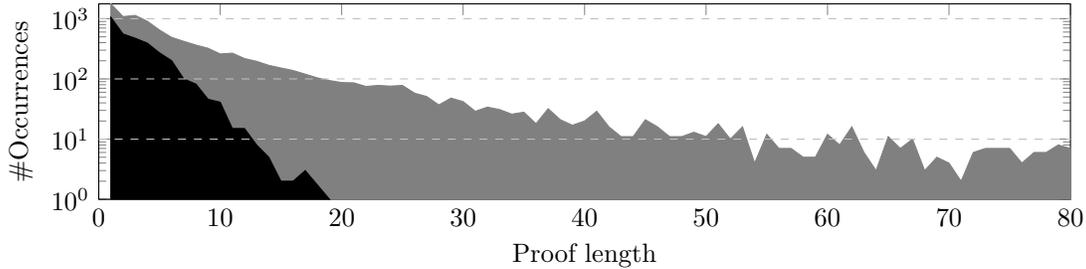

\Cref{fig:proof-lengths} compares the lengths of the proofs found by the LSHF
configuration with the lengths of the original proofs, measured in the number of
individual tactics executed. The system easily finds most of the short proofs.
Longer proofs are increasingly rare. However, a proof that utilizes 19
individual tactics was found. One striking thing to note is that the standard
library contains quite a few extremely long proofs (the graph cuts off at length
80, but a proof of length 1539 exists). These long proofs are not written
manually by the user but are the result of tactic sharing. An excellent example
of this is the \verb|ascii| inductive datatype. This type contains the obvious
256 cases. Proving a fact about \verb|ascii| using case analysis requires one to
prove all 256 cases individually. All cases can usually be proved using the same
tactic, for example, \verb|auto|. The human written proof then looks like
\verb|destruct c; auto|, while the system records 257 individual tactic
executions. Making use of such tactic sharing is future work related to the
issue of tactic recording granularity discussed in \Cref{sec:recording}.

\subsection{Comparison to CoqHammer}
CoqHammer utilizes several automatic theorem provers, such as
Vampire~\cite{DBLP:conf/cade/RiazanovV99}, E~\cite{DBLP:conf/lpar/Schulz13} and
Z3~\cite{DBLP:conf/tacas/MouraB08} with different settings and reconstruction
tactics to generate proofs. Individually, these strategies are reported to prove
between 17.5\% and 28.8\% of the Coq v8.5 standard
library~\cite{DBLP:journals/jar/CzajkaK18}. Combining all strategies leads to
solving 40.8\% library lemmas in total. Our individual configurations---and
especially the LSHF configuration that solves 34.0\% of the lemmas---are
generally better than CoqHammer's strategies, while the hammer narrowly beats us
in overall coverage.

More importantly, our system is quite complementary to CoqHammer. That is, we
can prove many different lemmas. A direct comparison between the two systems is
somewhat problematic due to the use of different versions of Coq's libraries.
The v8.5 library used for the CoqHammer evaluation has 9276 lemmas, while the
later v8.10 library used by us contains 10416 lemmas. Therefore, we compare only
lemmas that exist in both versions. A straightforward method for aligning the
lemmas in the two library versions yields 8231 common lemmas. Of those, 3240
(39.4\%) are solved by the union of our configurations, and 3534 (42.9\%) by the
union of the CoqHammer strategies. The union of both systems yields 4666
(56.7\%) proved lemmas, adding 32.0\% more solutions to those obtained by
CoqHammer. These results are a substantial improvement for Coq users who can
employ both methods and then expect over 55\% automation.

\section{Conclusion}
We have shown that machine learning for tactics is a useful enterprise. On its
own, our system can fully automatically prove 39.3\% of the standard library.
Together with CoqHammer, 56.7\% of the lemmas are proved, showing that these
approaches complement each other very well. Although the system is already quite
strong, many avenues of improvement exist. One can think of local parameter
prediction, incorporating local tactic history in predictions, implementing
better search strategies, and a proper user interface. We leave this as future
work.

\bibliographystyle{plainurl} \bibliography{bibliography}

\end{document}